\documentclass[letterpaper, 10 pt, conference]{ieeeconf}  

\IEEEoverridecommandlockouts                              

\usepackage{times} 
\usepackage{amsmath} 
\usepackage{amssymb}  
\usepackage{footnote}  

\usepackage{siunitx}

\usepackage{graphicx}

\usepackage{multirow}   

\usepackage{xcolor}

\usepackage[ruled, vlined, linesnumbered]{algorithm2e}
\SetAlFnt{\small} 
\SetAlCapFnt{\small} 
\SetAlCapNameFnt{\small} 

\usepackage[caption=false,font=footnotesize]{subfig}

\SetCommentSty{mycommfont}

\title{\LARGE \bf
Design for One, Deploy for Many:\\Navigating Tree Mazes with Multiple Agents
}

\usepackage{xcolor} 

\author{%
Jahir Argote-Gerald$^{1}$, 
Genki Miyauchi$^{1}$$^{\dagger}$, 
Julian Rau$^{2}$$^{\dagger}$,
Paul Trodden$^{1}$, 
and Roderich Gro\ss$^{1,2}$%
\thanks{$^{1}$ School of Electrical and Electronic Engineering,
The University of Sheffield, U.K.\,
{\tt\small \{jaargotegerald1, g.miyauchi, p.trodden\}@sheffield.ac.uk}}%
\thanks{$^{2}$ Department of Computer Science,
Technical University of Darmstadt, Germany.\,
{\tt\small \{julian.rau, roderich.gross\}@tu-darmstadt.de}}%
\thanks{$^{\dagger}$ Authors contributed equally to this work.}%
\thanks{$^{*}$This work was supported by IFARHU–SENACYT PhD scholarship, 
OpenSwarm project (EU’s Horizon Europe Framework Programme; grant 101093046), and
Robotics Institute Germany (BMBF; grant 16ME1001).}
}

\begin{document}

\maketitle
\thispagestyle{empty}
\pagestyle{empty}

\begin{abstract}

Maze-like environments, such as cave and pipe networks, pose unique challenges for multiple robots to coordinate, including 
communication constraints and congestion.
To address these challenges, we propose a distributed multi-agent maze traversal algorithm for environments 
that can be 
represented by 
acyclic graphs. It uses a leader-switching mechanism where one agent, assuming a head role, employs any single-agent maze solver while the other agents each choose an agent to follow. The head role gets transferred to neighboring agents where necessary, ensuring it follows the same path as a single agent would. 
The multi-agent maze traversal algorithm is evaluated in simulations with groups of up to 300 agents, various maze sizes, and multiple single-agent maze solvers. It is compared against strategies that are naïve, or assume either global communication or full knowledge of the environment. The algorithm outperforms the naïve strategy in terms of makespan and sum-of-fuel. It is superior to the global-communication strategy in terms of makespan but is inferior to it in terms of sum-of-fuel. 
The findings suggest it is asymptotically equivalent to the full-knowledge strategy with respect to either metric. 
Moreover, real-world experiments with up to 20 Pi-puck robots confirm the feasibility of the approach.

\end{abstract}

\section{INTRODUCTION}
\label{Intro}

The ability to navigate groups of robots through confined spaces 
is essential for 
real-world applications such as
pipe networks~\cite{parrott2020simulation}, caves~\cite{murphy2009mobile}, and other subterranean settings~\cite{martz2020survey}.
For example, in incidents such as the Tham Luang cave rescue~\cite{BeechPaddockSuhartono2018}, groups of robots 
could help
deliver supplies to a common, unknown location of victims.
Less evidently, human crowds can also create maze-like environments ~\cite{yang2019social,dugas2020ian}, 
where low permissivity levels result in human crowds behaving like unmovable walls. 

Many algorithms have been proposed to enable single robots to navigate maze-like environments. These single-agent maze solvers often represent the environments as graphs~\cite{cormen2022introduction,oleynikova2018sparse}.
Classic examples 
are depth-first search (DFS) and breadth-first search (BFS). They operate on arbitrary graphs, provided the robot has unbounded memory. Variants of DFS include wall-follower algorithms~\cite{alamri2023autonomous}, which guarantee the robot reaches any goal node, provided the maze is \emph{simple} (i.e., the graph representing it is a tree),
and Tr\'emaux's algorithm~\cite{even2011graph}, which also works on graphs with cycles by marking the entrance to passages at junctions it has visited.
However, these solvers do not consider multiple robots simultaneously exploring a common maze.

Various problems related to multi-agent
maze navigation 
have been studied. Some have used agents to collectively visit every location in a bounded environment~\cite{howard2002mobile}, 
similar to (uniform) coverage. For tree- or graph-like environments, 
agents 
have been tasked to visit every edge~\cite{fraigniaud2006collective,cabrera2012flooding} or node~\cite{dereniowski2015fast}. Others 
explored agents that collectively mapped graph-like environments~\cite{linardakis2024,linardakis2024multi}. 
These algorithms 
are effective for \emph{exploration} or \emph{mapping}, but do not address how groups should navigate towards a common, undisclosed goal.

Collision-free navigation is extensively explored in multi-agent path finding (MAPF) problems~\cite{silver2005cooperative,wagner2015subdimensional,sharon2015conflict,stern2019multi}. These require agents to move from their initial positions to designated goal positions. 
Agents typically have full knowledge of the environment, including a map and their start and goal positions.
In some studies, parts of the map must be discovered ~\cite{nebel2019implicitly,hall2020self,queffelec2023complexity,shofer2023multi}. For example, some edges are marked \emph{conditional}, and their state (e.g., open or closed) must be identified by visiting them. 

\begin{figure}[t]
    \centering

    \subfloat[]{
        \includegraphics[width=.3\columnwidth]{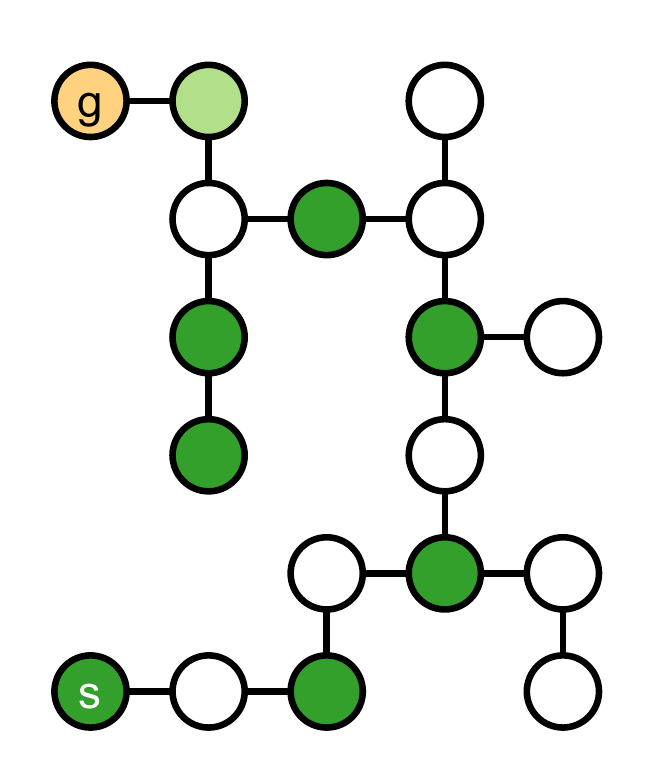}
        \label{fig:mazeenv}
    }
    \subfloat[]{
        \includegraphics[width=.3\columnwidth]{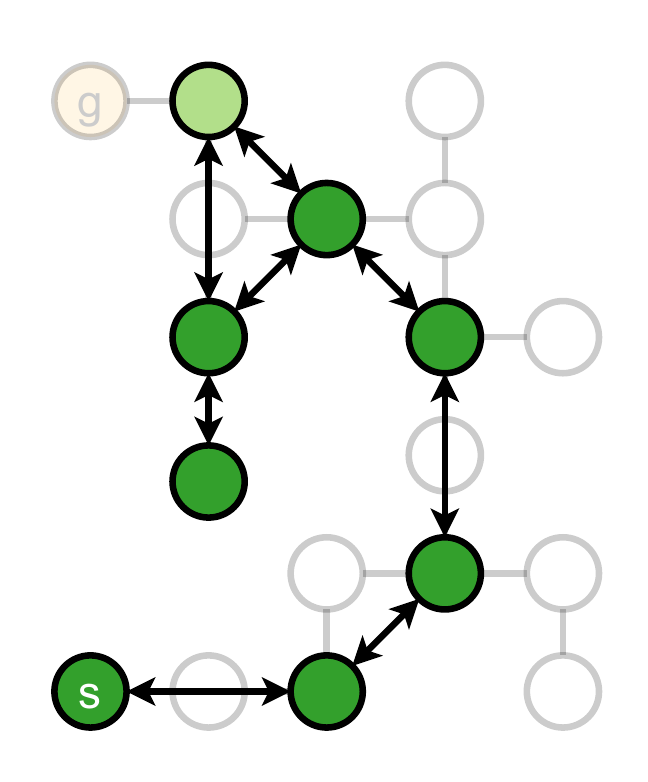}
        \label{fig:mazecoms}
    }
    \subfloat[]{
        \includegraphics[width=.3\columnwidth]{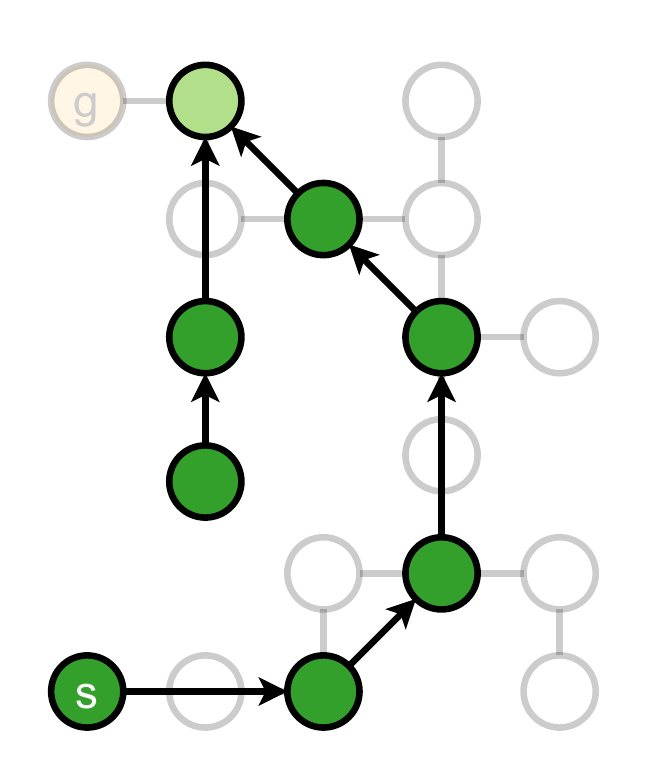}
        \label{fig::mazelead}
    }
    
    \caption{The multi-agent maze traversal problem. (a) Example maze represented as a connected, acyclic graph with start node $s$ and goal node $g$ (orange). Agents navigate the maze, with one agent, assuming the head role (light green), running a single-agent maze solver while all other agents (dark green) choose an agent to follow. The head role can transfer dynamically between neighboring agents. (b) Communication network between agents. (c) Tree graph representing the follower associations.}
    \label{fig:maze}
\end{figure}

This motivates research into a relatively unexplored problem we call \emph{multi-agent maze traversal} (MAMT). In MAMT, agents have no prior knowledge of the environment and must navigate to an undisclosed common goal, which could represent a maze exit. Such problems are relevant in the real world, where fleets of vehicles search for and collectively navigate to regions of interest in maze-like or densely populated environments.
Algorithms for solving MAMT problems in tree-like environments exist, but assume agents with centralized control~\cite{crnkovic2023fast} or global communication~\cite{kivelevitch2010multi}.

This paper proposes a distributed MAMT algorithm for navigating tree-like environments (Fig.~\ref{fig:maze}). A single agent becomes the leader of the group, referred to as the \emph{head}, and uses a given single-agent maze solver to explore the maze, while the other agents choose an agent to follow.
The head role gets dynamically transferred to neighboring agents where necessary.
We evaluate our approach with up to 300 simulated agents and 20 real robots
navigating mazes of various sizes. 
We also showcase how our approach compares against strategies that are naïve, or assume either global communication~\cite{kivelevitch2010multi} or full knowledge of the environment.




The paper is organized as follows. 
Section~\ref{PF} formulates the 
problem. Section~\ref{MT} presents an algorithmic solution.
Sections~\ref{SS} and \ref{RRE} present the results with simulated and real robots,
respectively.
Section~\ref{Conc} concludes the paper.

\section{PROBLEM FORMULATION}
\label{PF}

In this section, we introduce the 
MAMT problem considered in this work.
The maze is represented as a graph $\mathcal{G} = (\mathcal{V}, \mathcal{E})$ (Fig.~\ref{fig:maze}a). We assume $\mathcal{G}$ to be connected and acyclic, that is, a tree. Two nodes of this tree, labeled $s$ and $g$, represent the start and goal of the maze, respectively. 


The maze contains a set of $n$ agents, $\mathcal{A} = \{1, \dots , n\}$. Let $v_i[k]\in \mathcal{V}$ denote the node that agent $i\in\mathcal{A}$ resides on at time $k$.
For simplicity, we omit variable $k$ when it is clear from the context.
At time $k=0$, all agents are at the start of the maze (i.e., $\forall i: v_i[0]=s$). 

The status of node $u\in \mathcal{V}$ is defined as
\begin{equation*}
S^{\text{node}}_u =
\begin{cases}
\textit{occupied}, & \text{if }  u\neq g \wedge\exists i \in \mathcal{A}:
v_i = u \\
\textit{unoccupied}, & \text{otherwise}.
\end{cases}
\end{equation*}
The state of the goal is always unoccupied, thereby permitting all agents to move onto this node. 


We assume that agent $i\in \mathcal{A}$ knows
\begin{itemize}
\item its unique index (hereafter also \emph{ID}) within $\mathcal{A}$;
\item the node $v_i\in \mathcal{V}$ it currently resides on and whether
$v_i =s$, $v_i=g$ or $v_i \notin \{s,g\}$;
\item the set of nodes adjacent to the node at which it resides, $\mathcal{N}_i= \{ u\in \mathcal{V} \mid \{ v_i,u\} \in \mathcal{E}\}$; 
it can also determine a total order on $\mathcal{N}_i$, which is consistent across all agents;
\item the subset of adjacent nodes that are occupied, $\mathcal{N}_i^{\text{occupied}}= \{ u\in \mathcal{N}_i \mid S^{\text{node}}_u = \textit{occupied}\}$; we have $\mathcal{N}_i^{\text{unoccupied}}= \mathcal{N}_i \setminus \mathcal{N}_i^{\text{occupied}}$
\end{itemize}
and possesses any additional capabilities that the underlying single-agent maze solver relies on.

\begin{figure}
    \centering
    \includegraphics[width=0.45\linewidth]{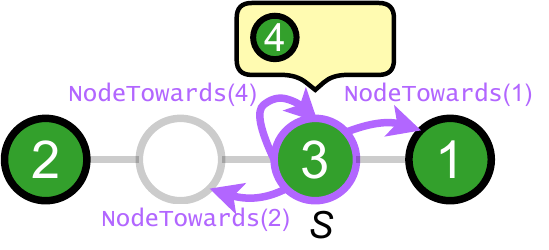}
    \caption{Illustration of the \texttt{NodeTowards}$_i(j)$ function from the perspective of agent $i=3$ at start node $s$. Purple arrows indicate the nodes from which agent 3 receives messages sent by neighboring agents $j$. 
    }
    \label{fig:nodewards}
\end{figure}

Each agent can communicate with other agents within its two-hop neighborhood, though messages cannot pass through occupied nodes. Formally, the communication graph is defined as $\mathcal{G}^{\text{com}}=(\mathcal{V}^{\text{com}},\mathcal{E}^{\text{com}})$, where 
$\mathcal{V}^{\text{com}}=\mathcal{A}$ and $\mathcal{E}^{\text{com}} = \{ \{i,j\}\subseteq\mathcal{A} \mid  i\neq j \land \left(v_i=v_j \lor \{ v_i,v_j\}\in \mathcal{E} \lor \exists u \in \mathcal{N}_i^{\text{unoccupied}} \cap \mathcal{N}_j^{\text{unoccupied}} \right)\}$.
Hence, agent $i$ communicates with agents in $\mathcal{C}_i = \{ j \in \mathcal{V}^{\text{com}} \backslash \{i\} \mid \{i,j\} \in \mathcal{E}^{\text{com}} \}$ (Fig.~\ref{fig:maze}b).
We assume that communication is \emph{situated}: where agent $i$ receives a message from agent $j$, the former knows the node from where the message reached itself, which is either the agent's node or a node adjacent to it. Hereafter this node is referred to as \texttt{NodeTowards}$_i(j)$ (Fig.~\ref{fig:nodewards}). Formally, 
\begin{equation*}
\texttt{NodeTowards}_i(j) =
\begin{cases}
v_j, 
& 
\begin{aligned}[t]
& \text{if } v_j = v_i \\
& \lor\, \{v_i,v_j\} \in \mathcal{E},
\end{aligned} 
\\
u \in \mathcal{N}^{\text{unoccupied}}_i, 
& \text{if } v_j \neq v_i \, \land \\ & u \in \mathcal{N}^{\text{unoccupied}}_j, 
\\
\text{undefined}, 
& \text{otherwise}.
\end{cases}
\end{equation*}

The objective is for all agents to reach goal node $g$. 
In time step $k+1$ an agent can decide either to move to any node $u \in \mathcal{N}^{\text{unoccupied}}_i[k]$, or to remain in its present node.
Moreover, at any time step, no two agents may reside on the same node of the graph, unless it is the start or goal node, and
no two agents may traverse the same edge, in the same or opposite direction.
In practice, the start and goal nodes could be considered staging zones that can each hold at least $n$ robots, such as 
an open area in front of, or a cavity inside, a cave~\cite{BeechPaddockSuhartono2018}.
An agent may move to the start node only when it is unoccupied.

We employ two
performance criteria
commonly used in MAPF~\cite{stern2019multi} problems:
(i) \textit{Makespan}: 
the total number of time steps required for all agents to reach the goal node;
(ii) \textit{average sum-of-fuel}: the mean distance moved per agent to reach the goal, reflecting per-agent energy use.

\section{MAMT ALGORITHM}
\label{MT}





\subsection{Algorithm Design}

The algorithm assumes the availability of a single-agent maze solver. At any time, one agent, hereafter the \emph{head} of the group, decides its next movement using the single-agent maze solver, whereas
all other agents choose a neighboring agent, hereafter their respective \emph{leader}, to follow, such that they are all direct or indirect followers of the head (Fig.~\ref{fig:maze}c). 
When executing the single-agent maze solver, the head agent does not take other agents into account:
If it decides to move towards a node already occupied by another agent, 
rather than moving itself, 
it transfers the head role to that agent.
This approach not only helps prevent collisions. It ensures that the head follows exactly the same path through the maze as a single agent running the single-agent maze solver would. 
Provided the single-agent maze solver finds the goal, our algorithm ensures that all agents reach the goal.

To enable coordination among agents, 
they exchange information with their neighbors. Agent $i$ receives the following information via messages from neighboring agents $j \in \mathcal{C}_i$:
 \begin{itemize}
     \item $j$'s ID and whether $v_j =s$, $v_j=g$ or $v_j \notin \{s,g\}$,
     \item $j$'s leader pointer $L_j$, and
     \item the node from which the message was received,\\  \texttt{NodeTowards}$_i(j)$.
 \end{itemize}

\begin{algorithm}[t]
\DontPrintSemicolon
\caption{Multi-agent Maze Traversal Algorithm}\label{mainalgorithm}

{\tcp{Initialize}} \label{initstart}
$v \gets s$\; \label{initatS}
Sense adjacent nodes\; \label{initsense}
Send and receive messages from neighbor agents\; \label{initcomms}
Update $\mathcal{C}$\;   \label{algo:comagentsinit}

$\mathcal{L} \gets \{\mathcal{C} \cup \{i\} \}$\; \label{setfirsthead}
\uIf{$ \texttt{Selector}(\mathcal{L}) = i$}{ \label{selectedasfirsthead}
$L \gets \text{nil}$ \label{initLnil} {\tcp*{\textcolor{gray}{Become head}}
}}
\Else{
$L \gets \texttt{Selector}(\mathcal{L})$ \label{initLnotnil} {\tcp*{\textcolor{gray}{Select leader}}}
}

Internally update leader of each agent in $\mathcal{C}$\; \label{inferagentsleaders}

\While{$v \ne g$}{
\label{rec1}
    {\tcp{Decision making step}}
$D \gets v$\; \label{initD}
\eIf{$L = \text{nil}$}{ \label{starthead}
    $D^{\text{solver}} \gets$ \texttt{SingleAgentMazeSolver}( )\; \label{solver}
        $\mathcal{H}  \gets \texttt{CompetingAgents}(\{D^{\text{solver}}\})$\; \label{headcandidates}
        
        \uIf{$\mathcal{H} = \emptyset$}{ \label{conditionpasshead}
            $D \gets D^{\text{solver}}$\; \label{movetoDsolver}
        }
        \Else{
            $L \gets \texttt{Selector}(\mathcal{H})$\; \label{newleadertoken}
            Prepare head transfer request to agent $L$\; \label{preparehtransfer}  
        }
} 
{
\label{startnonhead}
        $L^* \gets L$\; \label{labelcurrentleader}
        $L \gets \texttt{ResolveLeaderConflict}(L^*)$\; \label{checkconflict}
        $D_L \gets \texttt{NodeTowards(}L \texttt{)}$ \; \label{nodetowardsleader}
        
    \If{$ L^* = L \ \textbf{and} \ (v_L = g \ \textbf{or} \ D_L \notin \mathcal{N}^{\text{occupied}}) $}{ \label{conditionmovetoleader}
        $D \gets D_L$ \label{followertarget}\;
    }
}\label{endnonhead}

    {\tcp{Movement step}}
    \textbf{move to} $D$\; \label{move}
    {\tcp{Sensing and messaging step}}
    Sense adjacent nodes\; \label{sensealg1}
    Send and receive messages from neighbor agents\; \label{sendinfo}
    Update $\mathcal{C}$ \; \label{updateC}

    {\tcp{Select new leader}}
    \If {$L \notin \mathcal{C} \ \textbf{and} \ L \ne \text{nil}$} 
    {
    \label{healleadergraph}
    $L \gets \texttt{Selector}(\{ a \in \mathcal{C} \mid \texttt{NodeTowards}(a) = D_L\})$\; \label{selecthealedleader}
    }
    
    {\tcp{Head transfer step}}
    \If{Head transfer request received}{\label{ifreceivedhead}
        $L \gets$ nil\; \label{receivedhead}
    }
}
\end{algorithm}

\begin{algorithm}[t]
\DontPrintSemicolon
\caption{ResolveLeaderConflict}\label{alg:ResolveLeaderConflict}

\KwIn{A leading agent $L$}
$\mathcal{R} \gets \texttt{CompetingAgents}( \{\texttt{NodeTowards}(L), v\}  )$ \label{alg2:getR} \;
\uIf{$\texttt{Selector}(\text{$\mathcal{R} \cup \{i\}$}) = i$} { \label{alg2:conditiontokeep}
    \Return $L${\tcp*{\textcolor{gray}{keep leader}}}\label{alg2:keepL}
}
\Else{
    \Return $\texttt{Selector} (\text{$\mathcal{R} \cup \{i\}$})${\tcp*{\textcolor{gray}{update leader}}}\label{alg2:updateL}
}
\end{algorithm}

\begin{algorithm}[t]
\DontPrintSemicolon
\caption{CompetingAgents}\label{alg:CompetingAgents}

\KwIn{A set of nodes $\mathcal{U}$}
\Return $\{ a \in \mathcal{C} \mid (v_{L_a} \ne s \lor v_a \ne s) \land (a \ne L \lor \texttt{NodeTowards}(a) =v) \land \texttt{NodeTowards}(a) \in \mathcal{U} \land v_a \ne g \}$ \label{alg3:competingagents}

\end{algorithm}

The proposed MAMT algorithm is shown in Algorithm~\ref{mainalgorithm}.\footnote{Note that the $i$ subscripts are omitted since Algorithms~\ref{mainalgorithm}--\ref{alg:CompetingAgents} are presented from agent $i$'s perspective.}
Initially, each agent is at the start node $s$ and updates its knowledge of neighboring nodes and agents 
(lines \ref{initatS}--\ref{algo:comagentsinit}) at time $k=0$.
The agent determines whether to assume the head role in lines \ref{setfirsthead}--\ref{initLnotnil} by applying the \texttt{Selector} operator to the set of IDs of itself and all agents in its communication range (line \ref{setfirsthead}).
The \texttt{Selector} is a deterministic operator that chooses an ID from a set of robot IDs.
In this study, we use the $\min$ operator; therefore, \texttt{Selector} returns the lowest ID.
If the agent is chosen to be the head of the group, it sets its leader pointer $L$ to \text{nil} (line \ref{initLnil}). Otherwise, it sets its leader pointer to the ID of the head (line \ref{initLnotnil}). Then, it internally updates its knowledge about the leader pointer of other non-head agents to be the head (line \ref{inferagentsleaders}).

While not at goal $g$, the agent loops through
lines \ref{rec1}--\ref{receivedhead}. Time step $k$ increments at the end of every cycle.
The agent stores its current node $v$ in variable $D$ (line \ref{initD}).
The head agent ($L = \text{nil}$) determines whether to move to an adjacent node or transfer the head role (lines \ref{solver}--\ref{preparehtransfer}). 
It runs the \texttt{SingleAgentMazeSolver} function to determine the next node to move to 
and stores this as
$D^{\text{solver}}$ (line \ref{solver}).
Next, it uses the \texttt{CompetingAgents} function (Algorithm \ref{alg:CompetingAgents}) to determine $\mathcal{H}$, the set of agents within its communication range that are either residing on, or adjacent to, node $D^{\text{solver}}$. If multiple agents in its communication range are at the start node, only those agents that point to a leader outside of the start node are considered.\footnote{Agents at the goal do not have to be considered here, as the head agent is always the first to reach the goal, remains at this node, and does not transfer the head role.} If $\mathcal{H}$ is empty, the head agent can safely move to node $D^{\text{solver}}$ (line \ref{movetoDsolver}). Otherwise, 
at least one agent in $\mathcal{H}$ will necessarily reside on node $D^{\text{solver}}$ at the next time step.\footnote{As the graph is a tree, there exists a unique path between the head agent and any other node. All agents other than the head agent seek to move towards the head along their unique path.} 
The head transfers its role to the agent determined by \texttt{Selector} and remains stationary (lines \ref{newleadertoken}--\ref{preparehtransfer}).


A non-head agent ($L \ne \text{nil}$) seeks to move towards its leader (lines \ref{startnonhead}--\ref{endnonhead}).
It first checks whether the leader pointer must be updated (line \ref{checkconflict}).
The \texttt{ResolveLeaderConflict} function (Algorithm \ref{alg:ResolveLeaderConflict}) is used to resolve potential conflicts when multiple agents try to follow the same leader.
It checks if the agent should retain $L$ as its leader or switch to a different agent based on the \texttt{Selector} operator.
The competing agents are every agent $a \in \mathcal{C}$ that (i) either is not at the start node or their leader is not at the start node, (ii) is not agent $L$ (unless $L$ is in the same node as the current agent), (iii) can be reached via the same node as the leader (i.e., $\texttt{NodeTowards}(a)=\texttt{NodeTowards}(L)$) or is in the same node as the current agent, and (iv) is not at goal node $g$. 
If the \texttt{Selector} operator returns the agent's ID, the leader is to be kept. Otherwise, the leader is updated to be the selected agent. 
Subsequently, the agent retrieves the node towards leader $L$ (line \ref{nodetowardsleader}), which is stored in $D_L$.
The agent only uses $D_L$ as its target if the following conditions are met: (i) it kept its leader from the previous time step, and (ii) either its leader, $L$, resides on node $g$ or node $D_L$ is not occupied.

Next, the agent moves to $D$ (line \ref{move}), and 
then checks its adjacent nodes to communicate with its neighboring agents (lines \ref{sensealg1}--\ref{updateC}). If an agent's communication to their leader was blocked by another agent, it updates its leader to the blocking agent (line \ref{healleadergraph}--\ref{selecthealedleader}).
Finally, the agent checks 
if it received the head role from another agent (line \ref{receivedhead}).
Upon reaching the goal, it sends a final message to its neighbors before exiting the maze so that they can also reach the goal.

\begin{figure}[t]
    \centering

    \subfloat[$k=0$]{
        \includegraphics[width=.23\linewidth]{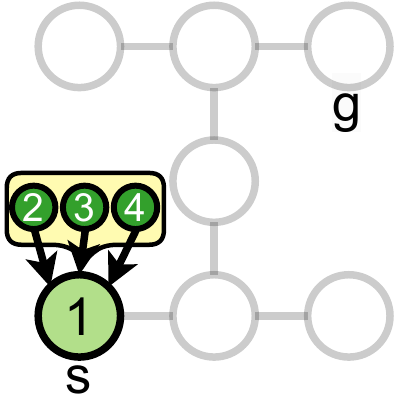}
    }
    \hfill
    \subfloat[$k=1$]{
        \includegraphics[width=.23\linewidth]{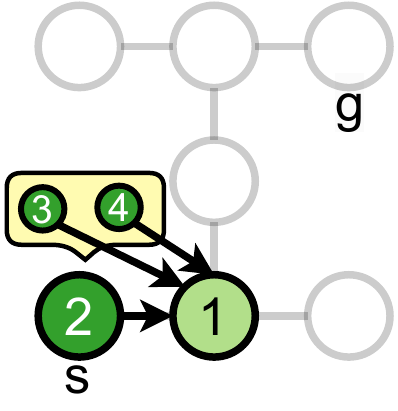}
    }
    \hfill
    \subfloat[$k=2$]{
        \includegraphics[width=.23\linewidth]{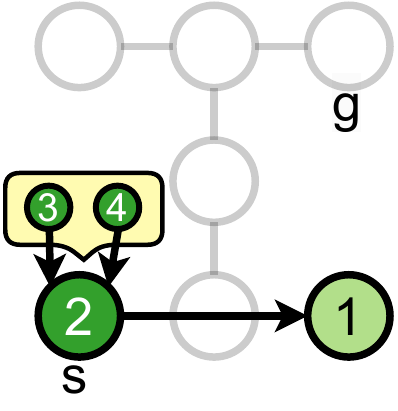}
    }
    \\
    \subfloat[$k=3$]{
        \includegraphics[width=.23\linewidth]{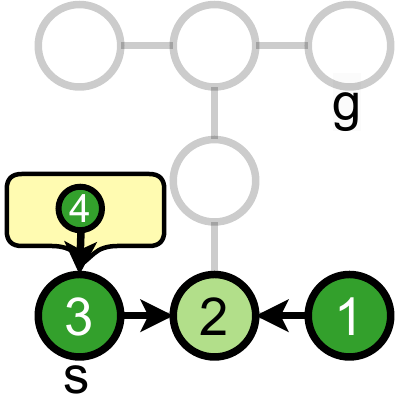}
    }
    \hfill
    \subfloat[$k=4$]{
        \includegraphics[width=.23\linewidth]{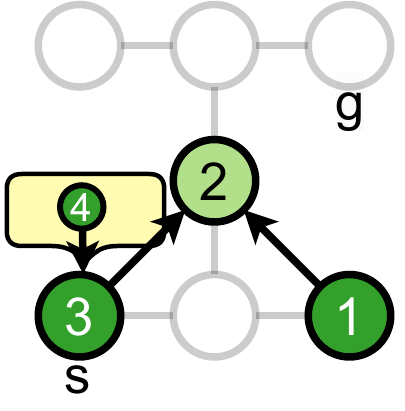}
    }
    \hfill
    \subfloat[$k=5$]{
        \includegraphics[width=.23\linewidth]{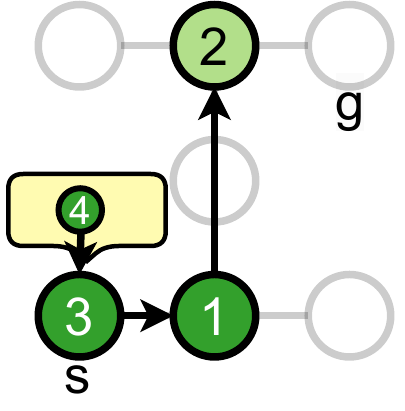}
    }    
    
    \caption{Example illustrating how the agents move and transfer the head role. 
    (a) All agents initialize at the start node, and one agent becomes the head agent (light green).  
    (b) The head moves to an adjacent node according to the single-agent maze solver, other agents (dark green) select agent 1 as their leader.  
    (c) The head explores the maze further.  
    (d) Agent 2 moves to follow the head, while the head role is transferred to agent 2.  
    (e) Two agents following the same leader compete for the same node.  
    (f) The followers resolve the conflict by selecting the smallest ID as the follower. The other agent updates its leader pointer.}
    \label{fig:snapshots}
\end{figure}

Fig.~\ref{fig:snapshots} illustrates 
how agents navigate the maze and transfer the head role.
At $k=0$, all agents are located at the start node (Fig.~\ref{fig:snapshots}a). The agent with the smallest ID becomes the head, and all others set their leader pointer to it.
The head moves to an adjacent node as determined by the single-agent maze solver; other agents remain at the start node (Fig.~\ref{fig:snapshots}b).
At $k=2$, the head reaches a dead end and decides to backtrack, but notices that the adjacent node will be occupied by agent 2 in the next timestep (Fig.~\ref{fig:snapshots}c). Hence, it chooses agent 2 as its leader and transfers the head role (Fig.~\ref{fig:snapshots}d).
The new head continues exploring the maze, while other agents sharing the same leader decide who remains the immediate follower and update their leader pointers accordingly (Figs.~\ref{fig:snapshots}e--f).

\subsection{Space Complexity}
Each agent requires $O(n)$ to store information about its neighbors (e.g., for leader selection at the start node) plus $O(1)$ for various variables (e.g., $L$ and $D$).
Additional memory required depends on the single-agent maze solver: wall-following DFS or (uniform) random walk needs only $O(1)$ of memory per agent (total $O(n)$), whereas BFS requires $O(|\mathcal{V}|)$ per agent (total $O(|\mathcal{V}|+n)$).
\subsection{Time Complexity}
The worst-case number of iterations required by Algorithm 1 depends on the single-agent maze solver:
wall-following DFS takes $O(|\mathcal{V}|)$ as $|\mathcal{E}|=|\mathcal{V}-1|$; BFS, which checks and backtracks at each node, takes $O(|\mathcal{V}|^{2})$; and a uniform random walk is expected to take $O(|\mathcal{V}|^2)$~\cite{motwani1996randomized}. Once the head reaches the goal, an additional $O(n)$ iterations are required for all agents to reach the goal. Each iteration takes 
$O(n)$ time, as the number of function calls is constant and each runs in $O(n)$ time.
Thus, the total time complexity is $O((|\mathcal{V}| + n)n)$ for DFS, $O((|\mathcal{V}|^2 + n)n)$ for BFS, and expected $O((|\mathcal{V}|^2 + n)n)$ for random walk.\footnote{
Assuming unlimited capacity at the start node, leader selection takes $O(n)$.
If agents instead enter the maze sequentially using a centrally imposed order, and letting  $\Delta(\mathcal{G})<|\mathcal{V}|$ denote the maximum degree of the maze, we obtain $O\left((|\mathcal{V}| + n) \min\left(\Delta(\mathcal{G}), n\right)\right)$ when using the wall-following DFS. In most practical applications, we expect $\Delta(\mathcal{G}) \ll |\mathcal{V}|$ and for magnitudes of $n$ where $O(n)$ becomes significant, also $\Delta(\mathcal{G}) \ll n$.}

\section{SIMULATIONS}
\label{SS}
\subsection{Simulation Setup}

We conduct simulations on a grid map consisting of connected nodes that represent navigable spaces, while blocked nodes represent walls.
Mazes of size 5\(\times\)5, 10\(\times\)10, 20\(\times\)20, and 30\(\times\)30 are generated using a randomized version of Prim's algorithm~\cite{prim1957shortest}.
The start and goal nodes are randomly selected from the set of navigable spaces.
Although we evaluate our approach in a grid-based environment, the underlying concepts are applicable to general graph structures.

We test 
three single-agent maze solvers:
\begin{enumerate}
    \item \emph{DFS}: The current head explores the leftmost unexplored node and backtracks upon reaching a dead end. It effectively realizes a wall-following approach. 
    \item \emph{BFS}: The current head visits all adjacent nodes and then progressively increases the depth of the search.
    \item \emph{Random walk}: The current head uniformly randomly selects an adjacent node to move to.
\end{enumerate}

We simulate $n\in\{1,5,25,50,100,200,300\}$ agents, with 20 trials per configuration.
Trials are terminated when any of the following occur: (i) all agents successfully reach the goal node; (ii) an agent collides with another agent; or (iii) an agent moves into a wall--whichever happens first.
Simulations were performed using a custom-built simulator. The source code is available in~\cite{argote-gerald_2025_source}.

\begin{figure}[tb]
    \centering
    \begingroup
    \setlength{\tabcolsep}{4pt} 
    \renewcommand{\arraystretch}{1} 
    \begin{tabular}{ccc}
    \includegraphics[width=0.28\columnwidth]{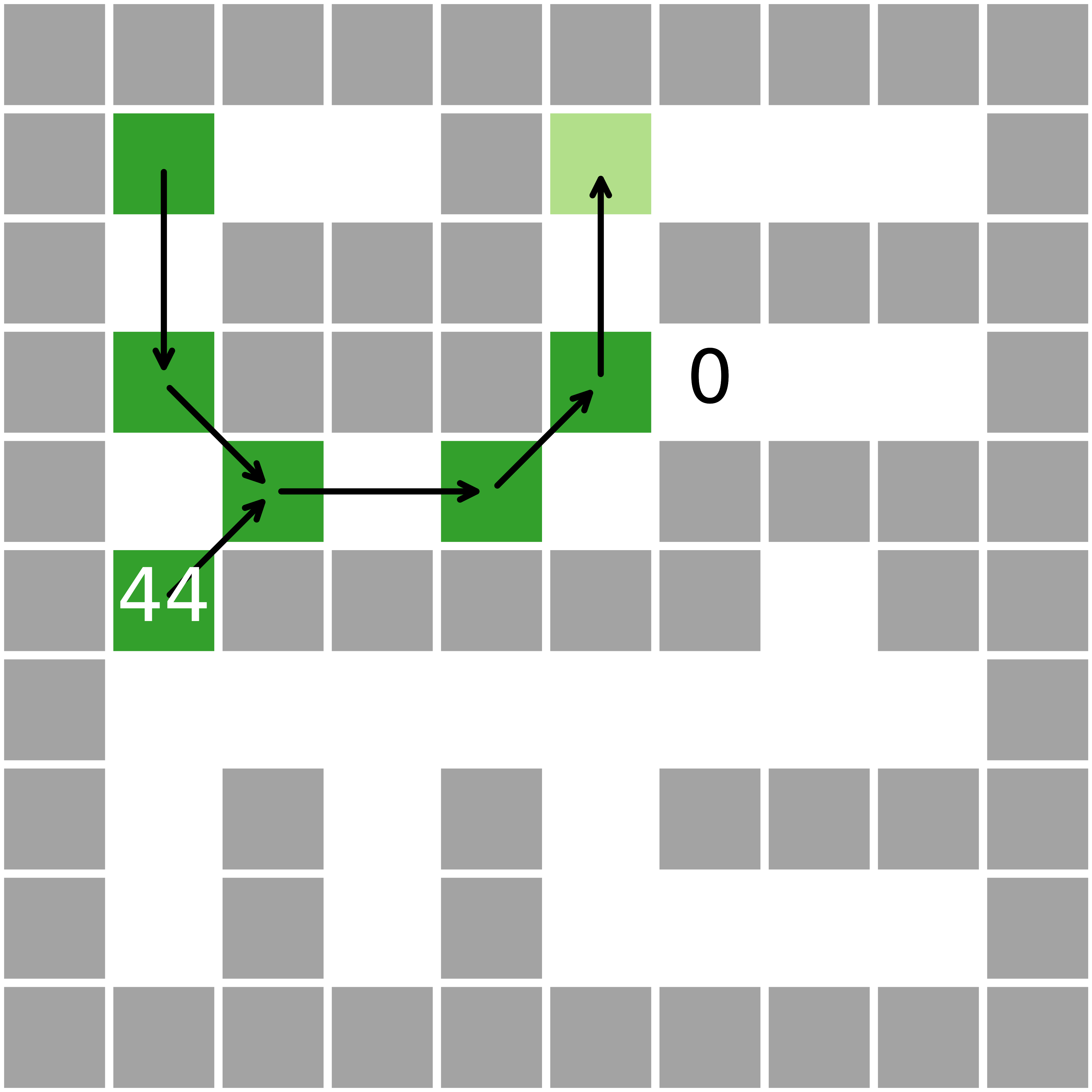} &
    \includegraphics[width=0.28\columnwidth]{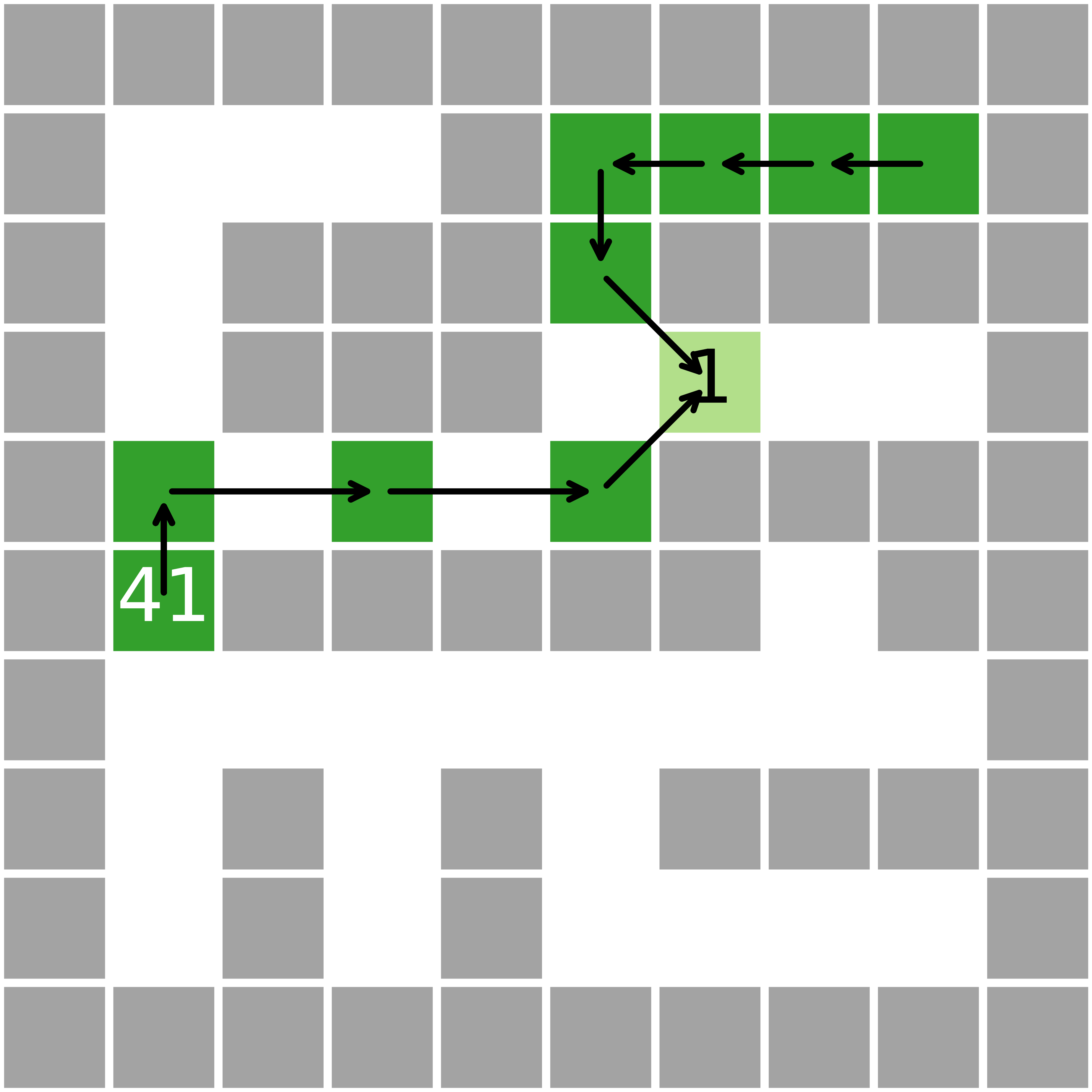} &
    \includegraphics[width=0.28\columnwidth]{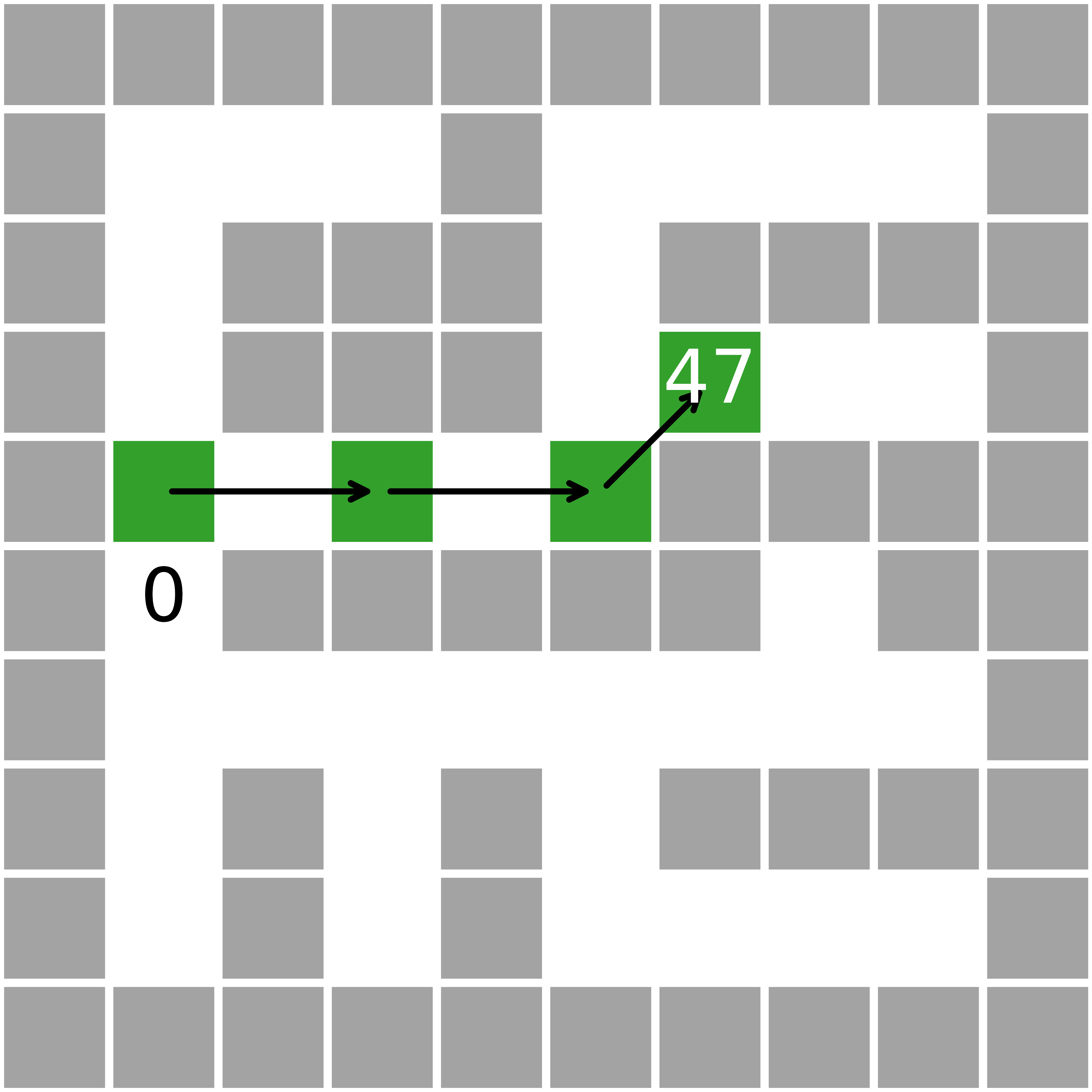}\\
    \small (a) $k = 18$ & \small \textbf(b) $k = 27$ & \small \textbf(c) $k = 119$\\

    
    \end{tabular}
    \endgroup
    \caption{Snapshots of 50 agents exploring a  10\(\times\)10 maze using 
    the DFS algorithm.
    The numbers on the left and right sides of the maze indicate the number of agents in the start and goal nodes, respectively.
    The head agent (light green) explores the maze, while the other agents (dark green) maintain a directed network to the head agent. 
    }
    \label{fig:screenshotexploration}
\end{figure}

\begin{figure*}[tb]
    \centering
    \begingroup
        \setlength{\tabcolsep}{0pt}
        \renewcommand{\arraystretch}{0}
        \begin{tabular}{ccc}
            \multicolumn{3}{c}{%
                \includegraphics[width=0.65\linewidth]{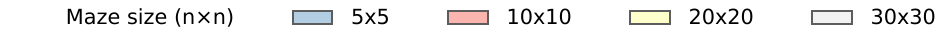}%
            } \\   
            \multicolumn{3}{c}{%
                \includegraphics[width=0.99\linewidth]{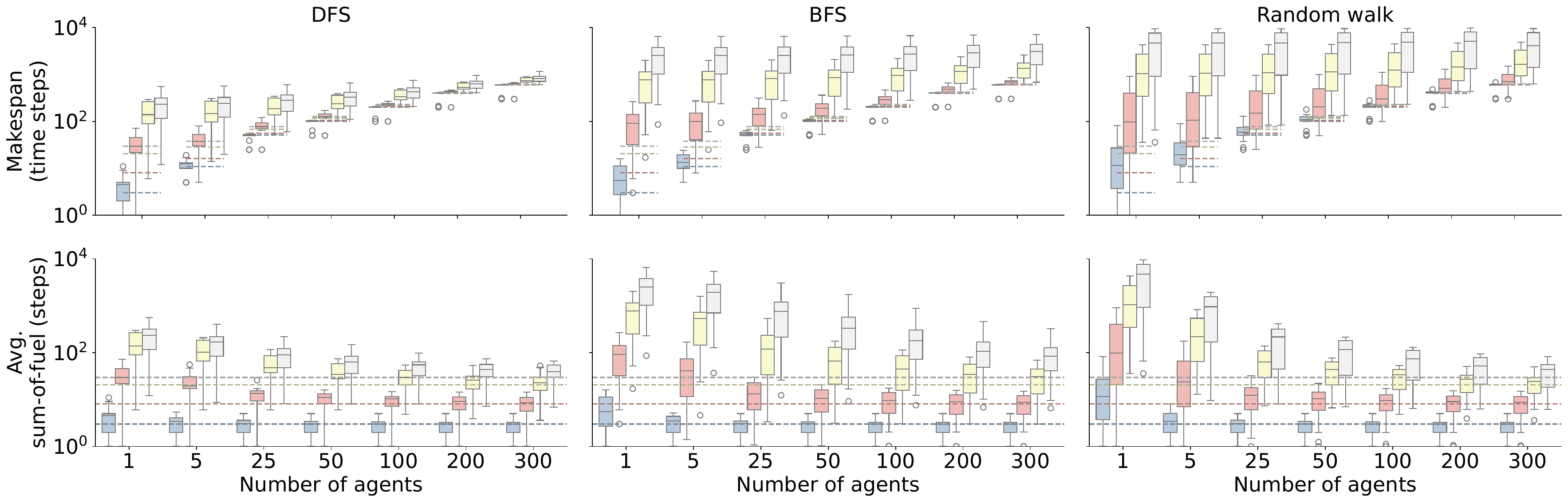}%
            } \\[4pt]   

        \end{tabular}
    \endgroup
    \caption{Makespan (top row) and average sum-of-fuel (bottom row) per agent for different maze sizes and numbers of agents using DFS, BFS, and random walk solvers.  
    Dashed lines indicate the median value for each maze size if agents followed the optimal path.  
    Each configuration was tested for 20 trials.}
    \label{fig:scalability_combined}
\end{figure*}

\subsection{Scalability Analysis}
\label{sec:scalability}




We evaluate the scalability of our proposed algorithm.
An additional timeout condition of 10,000 time steps was imposed to terminate trials in this analysis.
Fig.~\ref{fig:screenshotexploration} presents screenshots of 50 agents exploring a 10\(\times\)10 maze using the DFS 
maze solver.
The agents 
explore each branch in the maze one at a time, 
effectively implementing a wall-following algorithm.
Videos of trials for all single-agent maze solvers are available in the supplementary information.

Fig.~\ref{fig:scalability_combined} (top row) shows the makespan for different numbers of agents and maze sizes.
In general, larger mazes increase makespan due to longer travel distances.
Adding more agents also increases makespan as more time is needed for all agents to enter and exit the maze.
Once the head reaches the goal (Fig.~\ref{fig:screenshotexploration}b), the time it takes for the remaining agents to reach the goal is identical across all solvers.
This behavior results from the agents forming a connected tree, with each agent (except the head) following another agent while maintaining a one-node gap. 

Fig.~\ref{fig:scalability_combined} (bottom row) shows the average sum-of-fuel per agent.
As expected, larger mazes lead to longer average sum-of-fuel.
With more agents, the average sum-of-fuel approaches an asymptote, closely matching the sum-of-fuel of the shortest path between the start and goal nodes (shown as dashed lines). This suggests that, given enough agents, the average sum-of-fuel
approximates the optimal path length regardless of the maze size. Convergence is faster in smaller mazes due to fewer available routes.
Additionally, sum-of-fuel variability decreases as the number of agents increases, highlighting the benefit of deploying multiple agents.
Among the three solvers, DFS converges fastest and is the most effective.
Some random walk trials did not reach the goal before the timeout (21 trials for 20\(\times\)20 and 43 trials for 30\(\times\)30 mazes).
However, in all trials across solvers and group sizes, the agents never collided among themselves or with walls and the communication graph always remained connected, demonstrating the robustness of the MAMT algorithm and its compatibility with different single-agent maze solvers.

\subsection{Comparison with Naïve, Globally-Communicating, and Full Knowledge Agents}


We consider three alternative strategies:
\begin{enumerate}
    \item \emph{Naïve}: Agents independently run the single-agent maze solver. 
    Adjacent agents are treated as obstacles. If two agents are two nodes apart, the agent with the higher ID is given priority for moving to that node in between them, while the other treats it as an obstacle.
    \item \emph{Global-communication}~\cite{kivelevitch2010multi}: 
    Agents globally broadcast the nodes they visited and give higher preference to unexplored nodes during exploration.
    Once the goal has been found, all agents 
    move towards it directly.
    The approach relies on global communication, which assumes capabilities beyond those considered in our work. 
    To avoid collisions, the agents move sequentially, that is, only one agent moves at a time~\cite{kivelevitch2010multi}.
    \item \emph{Full knowledge}:
    Agents have access to a complete map of the environment (which is not available for agents in our work). 
    An agent moves towards the next node along the (unique) shortest path to the goal as soon as it becomes available.
\end{enumerate}
The MAMT algorithm and the naïve strategy used DFS, as it was shown to be the best-performing solver presented earlier. Trials were conducted in 20 different 20\(\times\)20 mazes.

\begin{figure}
    \centering
    \includegraphics[width=\columnwidth]{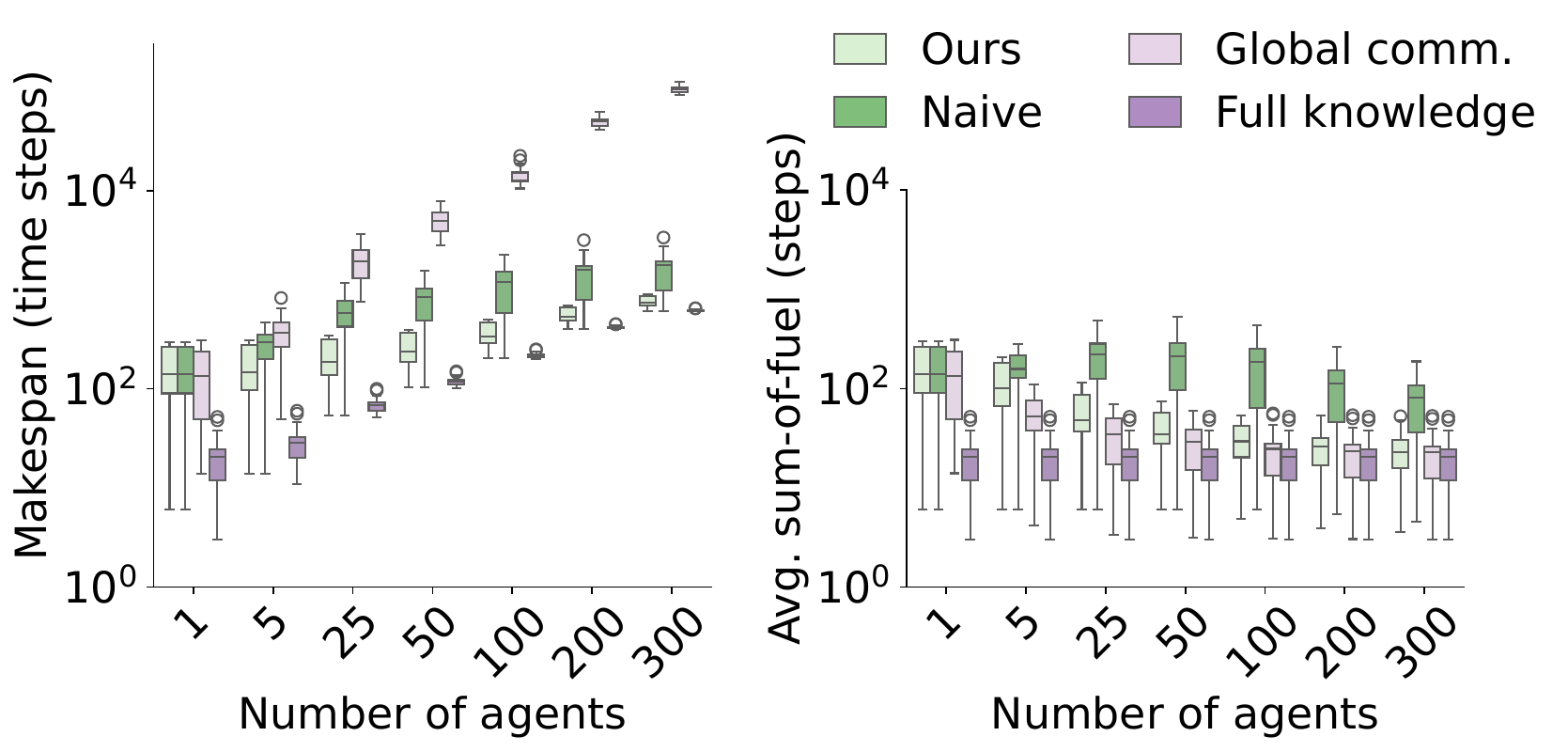}




    \caption{Comparison of our proposed approach against 
    the naïve, global communication~\cite{kivelevitch2010multi}, and full knowledge strategies regarding the makespan (left column) and the average sum-of-fuel per agent (right column). Agents used a DFS maze solver. Results are shown for 20\,$\times$\,20 mazes. Each configuration was tested for 20 trials.}
    \label{fig:uncoordplots}
\end{figure}

Fig.~\ref{fig:uncoordplots} (left column) shows the makespan.
The makespan naturally increases with the number of agents. The sharpest rise is observed for the global communication strategy, which prevents multiple agents from moving simultaneously. This is followed by the naïve strategy. The makespan of the MAMT algorithm exhibits only a moderate rise, asymptotically approximating the makespan of the full knowledge strategy.
These results highlight the scalability of our approach.


Fig.~\ref{fig:uncoordplots} (right column) shows the average sum-of-fuel per agent. 
Under the naïve strategy, the sum-of-fuel initially increases with more agents but starts decreasing beyond 50 agents.
In contrast, the MAMT algorithm consistently reduces
the average sum-of-fuel
as the number of agents increases; for $n=300$, the reduction amounts to 84.6\%. The global communication strategy achieves even lower sum-of-fuels, likely because agents explore multiple frontiers in parallel, increasing the chance of discovering the goal early.
The full knowledge strategy achieves the optimal sum-of-fuel, with all agents moving directly to the goal. As the number of agents increases, the sum-of-fuel of the MAMT algorithm approaches that of the full knowledge strategy.
\section{REAL ROBOT EXPERIMENTS}
\label{RRE}

\subsection{Experimental Setup}
We conducted experiments with up to 20 real Pi-pucks~\cite{millard2017pi} in a $2\,\si{\meter}\times2\,\si{\meter}$ arena featuring a virtual maze overlay.
To generate the virtual maze, the entrance and exit of the maze (i.e., start and goal nodes) were placed in the upper-left and lower-right corners of the arena, respectively. 
Additional nodes were randomly distributed without any overlap with existing nodes.
Initially, all nodes were fully connected. Edges were weighted to reflect the Euclidean distance between nodes. Afterwards, a minimal spanning tree was extracted.
An example maze is shown in Fig.~\ref{fig:maze_env}a. 
The robots were manually arranged by their IDs in an ascending order within the start node. This reflects the order in which they are chosen to leave that node. The robot with the head role used DFS as the single-agent maze-solver. To move between nodes of the maze, the robots used an artificial potential field~\cite{howard2002mobile}.
An overhead camera tracked each robot's position through unique ArUco markers attached on their tops, and emulated the localized communication via Wi-Fi~\cite{millard2018ardebug}.
A trial was considered successful if all robots reached the goal node within a common time period.

\subsection{Results}

\begin{figure}[t]
    \centering
    \begin{minipage}[t]{0.49\columnwidth}
        \vspace{0.76em}
        \centering
        \includegraphics[width=\linewidth]{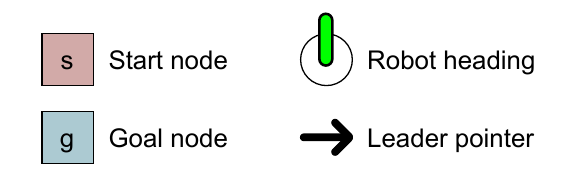}
        \vspace{1em}
        \subfloat[\label{fig:overall}]{
            \includegraphics[width=\linewidth]{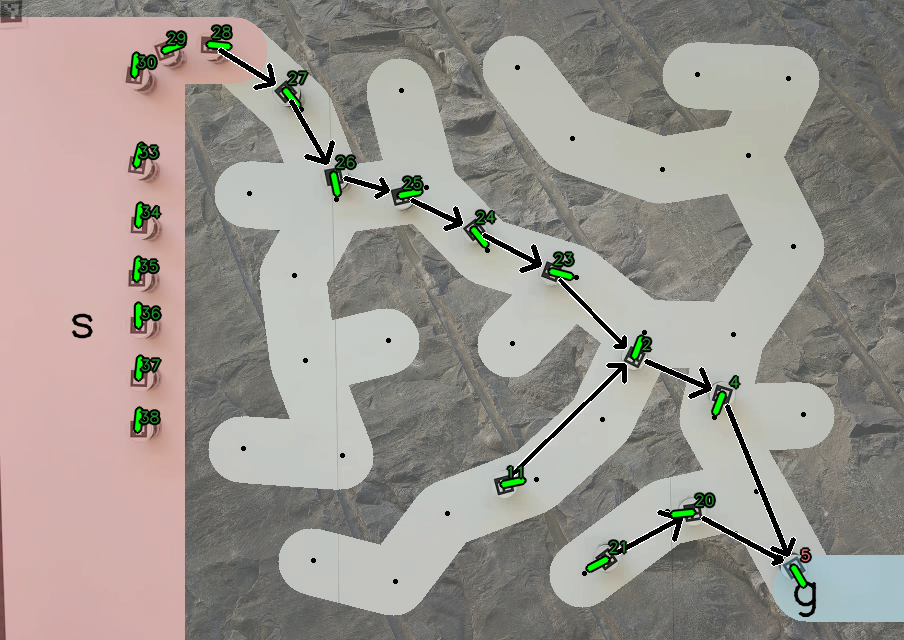}
        }
    \end{minipage}
    \hfill
    \begin{minipage}[t]{0.49\columnwidth}
        \vspace{0em}
        \centering
        \subfloat[\label{fig:detailA}]{
            \includegraphics[width=\linewidth]{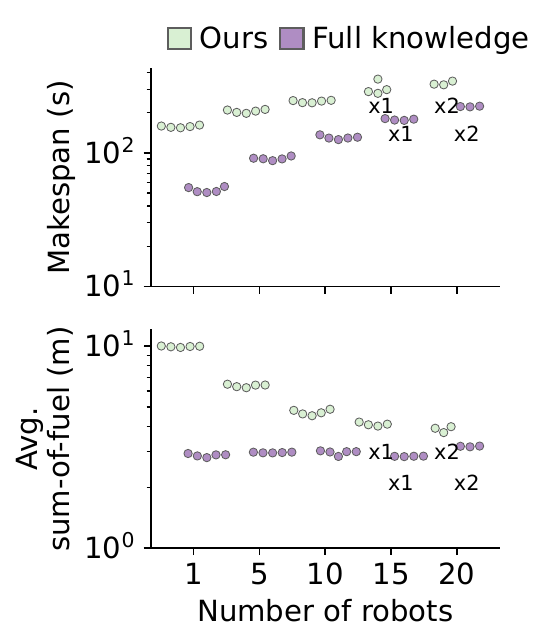}
        }
    \end{minipage}
\caption{Real-robot scalability study with Pi-puck robots. (a) Experimental arena: a virtual maze is overlaid on the workspace; robots start in the red zone and must reach the blue goal, with each graph-node center marked by a dot. For clarity, the arrows for the agents in the start node were omitted. 
(b) Makespan and average sum-of-fuel for groups of up to 20 robots running our proposed approach with a DFS maze solver and full knowledge robots.  
Each configuration was tested for 5 trials on the same maze. The number of failed trials is indicated for each configuration.}
    \label{fig:maze_env}
\end{figure}

For each $n \in \left\{1, 5, 10, 15, 20\right\}$ robots, we conducted five trials using both our approach and the full knowledge strategy. 
Fig.~\ref{fig:maze_env}b shows the results. Labels $\times1$ and $\times2$ indicate failed runs which were either due to losing Wi-Fi connection for 30\,s, battery depletion, or getting stuck due to surface unevenness.
The motion planner caused no deadlocks, livelocks, or rule violations.
As the number of robots increases, both the makespan and average sum-of-fuel of our approach converge towards those of the full knowledge strategy, suggesting asymptotically optimal behavior. 
This can be attributed to the facts that (i) DFS finds an exit in a finite time which does not depend on the number of robots, (ii)  once a robot discovers the goal, all robots remaining at the start node follow the shortest path---the same path that would be used by fully informed robots.

\section{CONCLUSION}
\label{Conc}

This work presented a fully distributed MAMT algorithm for navigating tree-like environments. 
One of the agents, called the head, runs a single-agent maze solver to identify its next move, whereas all other agents identify a neighbor that they follow to remain connected with the head. If the single-agent maze solver recommends the head to move towards other agents, rather than performing that move, the robot with the head role remains in its current node and requests the head role to be transferred to a suitable neighboring agent.

 
Through simulations and real-robot experiments across various maze sizes and numbers of agents, we analyzed the impact of different maze-solving algorithms
on navigation performance. Results showed that as the number of agents increased, the average sum-of-fuel decreased, approaching the optimal path length between the start and goal. 
This trend was observed for all examined single-agent maze solvers, although using DFS yielded the best results.

Future work will extend the approach to cyclic mazes, incorporating loop closure detection, and
enhance its robustness with respect to
unreliable communication.


\addtolength{\textheight}{-7cm}   





\section*{ACKNOWLEDGMENT}
We thank Mohamed S. Talamali for fruitful discussions about benchmarking the proposed coordination approach.




\bibliographystyle{IEEEtran}
\bibliography{main}

\end{document}